\documentclass{article}

\usepackage{arxiv}

\usepackage[utf8]{inputenc} % allow utf-8 input
\usepackage[T1]{fontenc}    % use 8-bit T1 fonts
\usepackage{hyperref}       % hyperlinks
\usepackage{url}            % simple URL typesetting
\usepackage{booktabs}       % professional-quality tables
\usepackage{amsfonts}       % blackboard math symbols
\usepackage{nicefrac}       % compact symbols for 1/2, etc.
\usepackage{microtype}      % microtypography

\usepackage{graphicx}
\usepackage[numbers]{natbib}
\usepackage{doi}

\usepackage{subfigure}
\usepackage{makecell} % for hyperparameter table

\title{Audiovisual Moments in Time: A Large-Scale Annotated Dataset of Audiovisual Actions}

\author{ \href{https://orcid.org/0000-0003-2175-0014}{\includegraphics[scale=0.06]{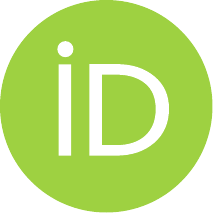}\hspace{1mm}Michael~Joannou}\thanks{Corresponding author} \\
	Computational Neuroscience and Cognitive Robotics Centre\\
	University of Birmingham\\
	Birmingham, UK \\
	\texttt{michaeljoannou@protonmail.com} \\
	\And
	\href{https://orcid.org/0000-0002-0301-9511}{\includegraphics[scale=0.06]{orcid.pdf}\hspace{1mm}Pia Rotshtein} \\
	University of Haifa\\
    Mount Carmel\\
	Haifa, Israel \\
	\texttt{pia.rotshtein@gmail.com} \\
    \AND
	\href{https://orcid.org/0000-0002-7697-2290}{\includegraphics[scale=0.06]{orcid.pdf}\hspace{1mm}Uta Noppeney} \\
	Donders Institute for Brain, Cognition and Behaviour\\
    Radboud University\\
	Nijmegen, The Netherlands \\
	\texttt{uta.noppeney@donders.ru.nl} \\
}

\date{}

\renewcommand{\headeright}{}
\renewcommand{\undertitle}{}
\renewcommand{\shorttitle}{}

\hypersetup{
pdftitle={Audiovisual Moments in Time},
pdfauthor={Michael ~Joannou}
}

\begin{document}
\maketitle

\begin{abstract}
We present Audiovisual Moments in Time (AVMIT), a large-scale dataset of audiovisual action events. In an extensive annotation task 11 participants labelled a subset of 3-second audiovisual videos from the Moments in Time dataset (MIT). For each trial, participants assessed whether the labelled audiovisual action event was present and whether it was the most prominent feature of the video. The dataset includes the annotation of 57,177 audiovisual videos, each independently evaluated by 3 of 11 trained participants. From this initial collection, we created a curated test set of 16 distinct action classes, with 60 videos each (960 videos). We also offer 2 sets of pre-computed audiovisual feature embeddings, using VGGish/YamNet for audio data and VGG16/EfficientNetB0 for visual data, thereby lowering the barrier to entry for audiovisual DNN research. We explored the advantages of AVMIT annotations and feature embeddings to improve performance on audiovisual event recognition. A series of 6 Recurrent Neural Networks (RNNs) were trained on either AVMIT-filtered audiovisual events or modality-agnostic events from MIT, and then tested on our audiovisual test set. In all RNNs, top 1 accuracy was increased by 2.71-5.94\% by training exclusively on audiovisual events, even outweighing a three-fold increase in training data. We anticipate that the newly annotated AVMIT dataset will serve as a valuable resource for research and comparative experiments involving computational models and human participants, specifically when addressing research questions where audiovisual correspondence is of critical importance.

\end{abstract}

\freefootnote{Preprint. Under review.}
\freefootnote{AVMIT is available at \url{https://zenodo.org/record/8253350}}

\section{Introduction}

% AV is important, and has been used in speech recognition
Many events generate auditory and visual signals that evolve dynamically over time. To obtain a more robust and reliable percept of the environment human observers integrate redundant and complementary information across sensory modalities \cite{NoppeneyReview2021}. For instance, audiovisual integration facilitates speech comprehension in noisy and adverse environments \cite{LeeNoppeney2011b}. As work in the area of deep learning has progressed, researchers have looked to take advantage of additional information available across multiple modalities to improve recognition performance. In speech recognition, for instance, researchers have developed deep neural networks (DNNs) to leverage audiovisual correspondences \cite{StavrosPetridis2017, Afouras2018}. To solve audiovisual speech recognition, DNNs rely on large labelled datasets with high levels of audiovisual correspondence \cite{Afouras2018, LRS3-TED}.

% Why was re-annotation necessary? There isn't a wealth of AV action event data
In the domain of action recognition, audiovisual events produce corresponding audio and visual signals, and these correspondences could be used to improve recognition rates \cite{Noppeney2010}. Despite the improved recognition rates available, annotations for the most popular large action recognition datasets are either visual-only or modality-agnostic (occurring in either/both modalities) \cite{Heilbron2015, Gu2018, Monfort2019, Li2020, Smaira2020}. This leads to a lack of audiovisual correspondence in available datasets, as an event may have only occurred in a single modality, or the auditory and visual signals may have accurately represented the labelled action despite being generated by different events.

% There are some options for AV action event data, but they aren't ideal
Although the majority of action recognition datasets are not annotated for audiovisual events (an event with both an auditory and visual signal) \cite{Heilbron2015, Gu2018, Monfort2019, Li2020, Smaira2020}, some researchers have begun to target the audiovisual domain in their data collection/annotation. \cite{Zhou2018} carried out a large-scale annotation task that assessed whether an event is present in both the audio and visual streams. But this annotation scheme only ensured that audio and visual signals corresponded to the label, not that they were caused by the same event. Another audiovisual action recognition dataset is Epic-Kitchens \cite{EpicKitchens2021} with videos depicting egocentric (1\textsuperscript{st} person) hand object interactions in kitchens. But the deep learning community still lacks a high quality allocentric (3\textsuperscript{rd} person) audiovisual action dataset.

% We present...
To facilitate deep learning research in the audiovisual domain, we present Audiovisual Moments in Time (AVMIT), a set of 57,177 audiovisual annotations for the Moments in Time dataset (MIT) \cite{Monfort2019}. To obtain AVMIT, we take a subset of the MIT dataset and run a large-scale annotation regime. Growing research reveals noncompliance \cite{Webb2022, Dennis2020} of participants on Amazon Mechanical Turk \cite{Crowston2012}, including \cite{Kell2018} were 49\% of turkers were found not to be wearing headphones despite reporting they did. To ensure high quality annotations, we elected to train raters and have them perform the task in a controlled lab setting. AVMIT contains 3 independent participant ratings for 57,177 videos (171,630 annotations). We further screened MIT videos to select a highly controlled audiovisual test set of 960 videos across 16 action classes, named the AVMIT test set. The AVMIT test set is suitable for human and DNN experimentation, particularly for studies concerned with audiovisual correspondence. Finally, to lower the computational requirements to train DNNs on audiovisual problems, we provide two sets of audiovisual embeddings that can be used to further train audiovisual DNNs. To obtain each set of audiovisual embeddings, we use convolutional neural networks (CNNs); VGGish \cite{Hershey2017} (audio) and VGG-16 \cite{Simonyan2015} (visual) or YamNet \cite{YamNet} (audio) and EfficientNetB0 \cite{Tan2019} (visual) and extract features from all AVMIT annotated videos.

% Additional uses
Beyond building audiovisual recognition models, AVMIT can be used for audiovisual separation (separating sounds from different sources), localisation (finding the sound source in the visual context), correspondence learning (discerning if the audio and visual signal emanated from the same source/type of source) and generation (generating audio from visual or visual from audio). Further, AVMIT serves as a valuable resource for research and comparative experiments involving computational models and human observers that are known to rely on audiovisual correspondences \cite{NoppeneyReview2021}. As DNNs are now commonly used as predictive models of human behaviour in vision \cite{Cichy2019} and audition \cite{Kell2018}, AVMIT supports this research to take a step into the audiovisual domain.

\section*{Methods}

\subsection*{Participants}
\label{subsec:participants}
To rate the videos, eleven participants (10 females; mean age 26.18, range 19-63 years) were recruited and gave informed consent to take part in the experiment. No participants were excluded. Each participant annotated a subset of the candidate videos. All reported normal hearing and normal or corrected-to-normal vision. Participants were reimbursed for their participation in the task at a rate of £6 per hour, plus a bonus of 10p paid for correct classification of randomly interspersed ground truths (further detailed in the Bonus Section).
Participants on average earned a total (hourly payment + bonus) of less than £7 per hour. The research was approved by the University of Birmingham Ethical Review Committee.

\subsection*{Experiment Setup}
\label{subsec:setup}
Participants were seated at a desk in an experiment cubicle or quiet area to complete this task. The experiment was presented on a Dell Latitude 5580 laptop with 15.6” screen and Linux Ubuntu 18.04.2 LTS operating system. Auditory stimuli were presented via a pair of Sennheiser HD 280 Professional over-ear headphones. The experiment was programmed in Python 2 \cite{van1995python} and Psychopy 2020.2.10 \cite{JonathanPeirce2019}.

\subsection*{Selection of MIT Videos}
\label{subsec:stimuli}
Prior to the annotation task, we carried out a selection process to obtain a subset of MIT videos that were more likely to contain audiovisual actions. We first obtained the labelled training (802,264 videos) and validation (33,900 videos) sets of the MIT dataset. The events depicted in these videos unfold over 3 seconds. For many of the classes in the MIT dataset, audio data would not help recognition of the labelled event (e.g. “imitating”, “knitting”, “measuring”). We carefully curated a subset of 41 audiovisual classes (corresponding to 88,579 training videos and 4,100 validation videos) that offer a wealth of informative audio and visual correspondences, enabling enhanced classification through the integration of these signals. These action classes are listed in the Figure \ref{fig:perceived and dominant AVMIT}.

\begin{table}[h]
\centering

\caption{\label{tab:bolstering list}Excluded MIT classes that were relabelled and added to the annotation task.}
\begin{tabular}{cc}
\hline
AVMIT class & Additional MIT class \\
\hline
Giggling & Laughing \\

Frying & Cooking, Boiling \\

Inflating & Blowing \\

Pouring & Spilling, Drenching, Filling \\

Diving & Swimming, Splashing \\

Raining & Dripping \\
\hline
\end{tabular}
\end{table}

To increase the number of videos in our selected AVMIT classes, we obtained videos from similar, but excluded, MIT classes, relabelled them, and added them to our annotation task. Incorrectly relabelled videos would be annotated by our participants as not containing the labelled audiovisual event. Table \ref{tab:bolstering list} displays those AVMIT classes alongside the other MIT classes that were relabelled and added to the annotation task. To ensure that candidate videos included audio and video components, we removed videos without audio streams or whose amplitude did not exceed 0 (digital silence).

\subsection*{Annotation Procedure}
\label{subsec:procedure}
Next, we created a video annotation task that could be carried out by multiple trained participants to identify if videos contained the labelled audiovisual event and whether it was the most prominent feature. This procedure was similar to the annotation procedure carried out in \cite{Zhou2018} to produce the VEGAS dataset.

Participants were presented with a series of audiovisual videos and were instructed to provide a button response after each had finished playing. On each trial, participants were presented with a 3 second video and then classified it as 1:“unclean”, 2:“moderately clean” or 3:“very clean”. To provide a classification, participants were trained to use the following logic:

\begin{enumerate}
    \item Was the labelled audiovisual event present?:
        
        No: give a 1 rating
        
        Yes: move to the next question
        
    \item Was the labelled audiovisual event the most prominent feature?:
        
        No: give a 2 rating
        
        Yes: give a 3 rating
        
\end{enumerate}

For this task, an event was considered to be the most prominent feature if it was of longer duration and higher intensity than any other event in the same video. Intensity related to amplitude of event audio and size of the event's region of interest. Each video was rated by at least 3 participants.

\begin{figure}[]
    \centering
        \includegraphics[width=1.0\linewidth]{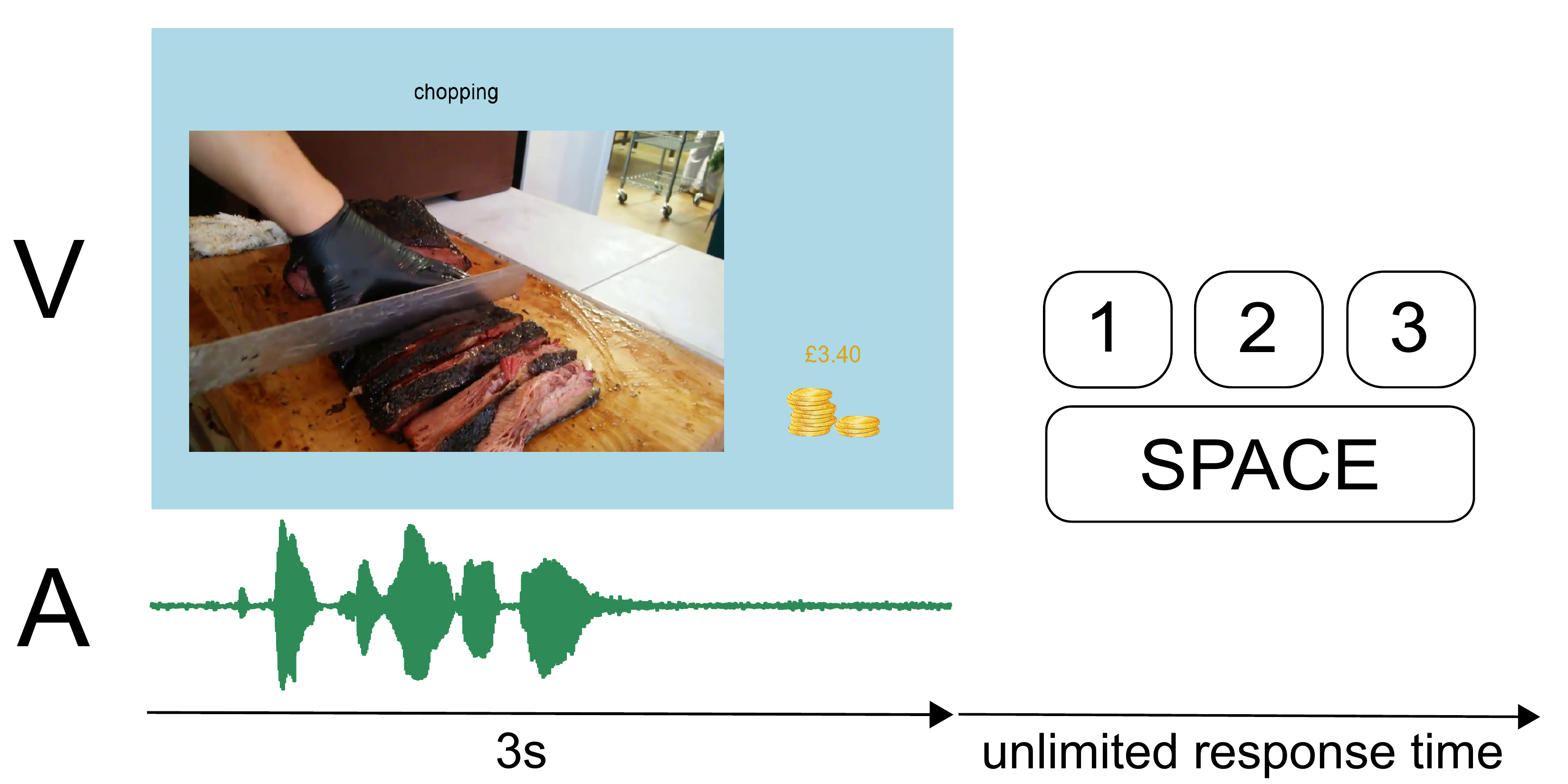}
    \caption{Annotation task schematic. Task screen displays a chopping video with label and accumulated bonus. Video plays for 3 seconds alongside audio stimuli. Participants watched and listened to the audiovisual video before providing a rating.}
    \label{fig:AVMIT sorting screen}
\end{figure}

During video presentation, the screen displayed the suggested action label at the top, the video in the bottom-left (videos had different resolutions so they were each given a common left edge position and bottom edge position) and a bonus counter in the bottom right (Figure \ref{fig:AVMIT sorting screen}). Together with the video, participants were presented with the audio via headphones. After the video and audio stopped playing, the program waited until the participant pressed a key. The options were; 1, 2, 3, space, where the numbers referred to the classification system described above and the space key would replay the video. Participants were able to replay the video and audio any number of times they like before making a classification. If the participant made a classification while the video was still playing, a warning screen would fill the display, instructing the participant not to press a key too early. This was particularly important given that the audiovisual video content after an early classification may change the answer to question 2. After a classification was made, the bonus counter would be updated, and the new label title and audiovisual video would appear.

\subsection*{Quality Control}
% SCREENING AND TRAINING
In order to ensure the quality of the AVMIT dataset, we opted to use trained participants in a controlled environment rather than Amazon Mechanical Turk. Participants were required to complete the training exercise, before they could participate in the annotation task. Before starting, each participant was given a set of instructions that outlined the task on a sheet of paper. These instructions were then verbally explained to them. The participants then undertook a training exercise whereby a video from each class was presented and the possible classification and reasoning was discussed with the author (MJ) of the study. The participants were then screened to ensure that they understood the task by classifying another set of videos (1 video per class) under the observation of the author. Of these videos, the participants needed to classify 38 of the 41 videos according to the author’s ground truth. Of the 11 participants that completed the training and testing exercise, all participants passed and went on to take part in the annotation task.

% BONUS PAYMENTS
Another strategy we employed, was to provide bonus payments to participants in order to ensure engagement and provide positive feedback. A bonus payment of 10p (GBP) was given for each classification of a video for which a ground truth was available. To obtain ground truths, 2,000 videos were uniformly sampled from the set of candidate videos prior to the annotation task and then classified by one of the authors (MJ). These audiovisual videos were distributed throughout the annotation task and participants were unaware of the possibility of a bonus when completing a trial. If the participant gave a matching classification for one of these previously classified audiovisual videos, they would receive a bonus, which was added to their total in the bottom right of the screen (Figure \ref{fig:AVMIT sorting screen}). This bonus accumulated over their sessions and was paid at the end of participation alongside their hourly compensation.

Quality of annotations was further ensured by using at least 3 participants to rate each video, in line with the procedure of other large dataset annotation schemes \cite{Zhou2018, Monfort2019, Li2020}. As the AVMIT annotation scheme was run using videos from an existing dataset, AVMIT benefits from the quality assurances of two cleaning processes.

\subsection*{Test Set}
% We obtained a highly controllable test set.
We ran further screening to obtain a highly controllable test set for human and deep neural network experiments. This process was 2 stages; class filtering and video filtering. Many classes did not contain a sufficient number of clean audiovisual videos for training and testing a deep neural network (Figure \ref{fig:perceived and dominant AVMIT}). We used a majority vote criteria to obtain those videos containing the labelled audiovisual event as a prominent feature. Classes with 500 or more videos that meet this criteria were accepted into the test set. Just 16 of 41 classes met this criteria, although this is in line with test sets in the humans vs. DNN literature \cite{Geirhos2018}. With test classes obtained, we then applied video filtering. In order to ensure reliability, we set as a criterion that all participants must agree that the  audiovisual event was present and the key feature of the video. In order to ensure a level of homogeneity in the dataset, we obtained those audiovisual videos with a visual frame rate of 30fps and further cleaned them, removing videos that:
\begin{itemize}
    \item Had been edited to appear as though something supernatural had occurred (such as something appearing or disappearing instantaneously)
    \item Had an excessive number of time-lapses
    \item Contained frames with excessive watermarks or writing on the frames
    \item Consisted of 2 video streams
    \item Were not naturalistic (depicting cartoons or simulations)
\end{itemize}

From the subset of filtered videos, 60 videos were uniformly sampled from each class and used to provide the AVMIT test set (60 videos per class, 16 classes, 960 video test set). By comparison, naturalistic stimuli sets for human experiments in the area of psychology and neuroscience often have far fewer stimuli \cite{Noppeney2010, ChanVideoStimuli, MichelleToMovieClipsPaper} and these may be further manipulated according to a variety of conditions to effectively multiply test set size. After filtering train videos with AVMIT in our experiments, this test set formed approximately 12\% of our total samples.

\subsection*{Neural Network Embeddings}
\label{subsec:embeddings}
We created 2 sets of audiovisual embeddings; those obtained using VGGish \cite{Hershey2017} and VGG-16 \cite{Simonyan2015} and a second set obtained using YamNet \cite{YamNet} and EfficientNetB0 \cite{Tan2019}. Both VGG-16 and EfficientNetB0 were trained on ImageNet \cite{Deng2009} and VGGish and YamNet were trained on AudioSet \cite{Gemmeke2017}. Prior to feature extraction by these CNN models, audio and visual data was preprocessed.

If the audio was stereophonic rather than monophonic, a monophonic stream was obtained using pydub.AudioSegment.set\_channels() \cite{Pydub2018}, taking the mean of the left and right channels (Equation \ref{eq:stereo to mono}). Where $S_{new}$ is the new monophonic audio sample, $S_L$ is the original left sample and $S_R$ is the original right sample.

\begin{equation}
\label{eq:stereo to mono}
    S_{new} = 0.5\cdot S_L + 0.5\cdot S_R
\end{equation}

% 16 bit, scaled and resampled
Audio data of a depth other than 16 bits was cast to 16 bits using \sloppy pydub.AudioSegment.set\_sample\_width() \cite{Pydub2018}. These int16 audio samples were then mapped from the range [-32768, 32767] ($2^{15}$ with one bit dedicated to sign) to the range [-1.0, 1.0] by dividing by the maximum value of 32768.0. The audio was then resampled to 16 kHz before spectrograms were calculated.

% Spectrogram
Next we carried out a short-time Fourier transform (STFT) to provide a frequency decomposition over time. We used a frame size of 25ms (the period over which signals are assumed to be stationary) and a 10ms stride (the frequency with which we obtain a frame). Overlapping frames help to ensure that any frequency in the signal that may exist between otherwise non-overlapping frames are captured in the spectrum. A Hann filter was applied to each of the frames before a fast Fourier transform (FFT) was carried out. A log mel spectrogram was then obtained using a mel filter bank of 64 filters, over the range 125-7500 Hz, and then finding the logarithm of each spectrum (plus a small delta of 0.01 to avoid taking the log of 0; Equation \ref{eq:log mel}).

\begin{equation}
\label{eq:log mel}
    log\ mel\ spectrogram = log(mel\ spectrogram + 0.01)
\end{equation}

% Windowing the spectrograms
The log mel spectrograms were windowed into smaller 960ms spectrograms, ready for the CNN. Audio preprocessing deviated between the VGGish and YamNet embeddings in this final stage of preprocessing in accordance with their training regimes \cite{Hershey2017, YamNet}. For VGGish, the stride was 960ms between windows, for YamNet, the stride was 480ms.

% Visual preprocessing
For visual processing, we sampled frames according to the frequency of the complementary audio features; 960ms for VGG-16 and 480ms for EfficientNetB0. This was to provide a similar number of audio and visual embeddings per sample. Frames were then resized to dimensions of 224x224x3 using OpenCV \cite{Opencv2000} in line with the expected input size of the CNN models. For VGGish the images were then zero centred, but for EfficientNetB0, images were rescaled, normalised and then zero-padded.

\section*{Data Description}
AVMIT consists of 4 components; audiovisual annotations of 57,177 MIT videos, a selection of 960 MIT videos designated as the AVMIT test set and 2 sets of audiovisual feature embeddings. All of these are available at \url{https://zenodo.org/record/8253350}.

\begin{figure}[h]
    \centering
    \begin{subfigure}
        \noindent
        \includegraphics[width=0.48\linewidth]{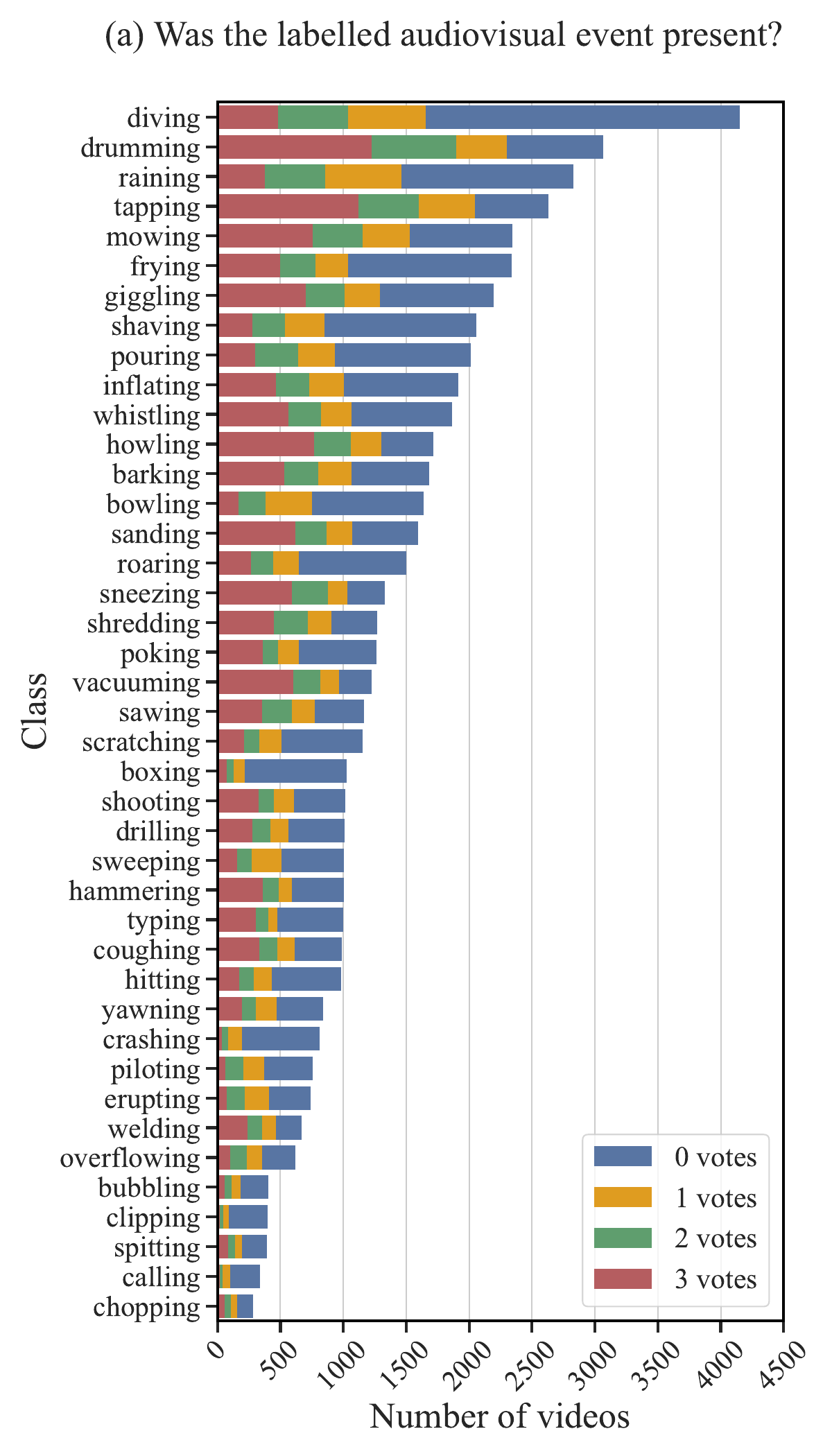}
    \end{subfigure}
    \begin{subfigure}
        \noindent
        \includegraphics[width=0.48\linewidth]{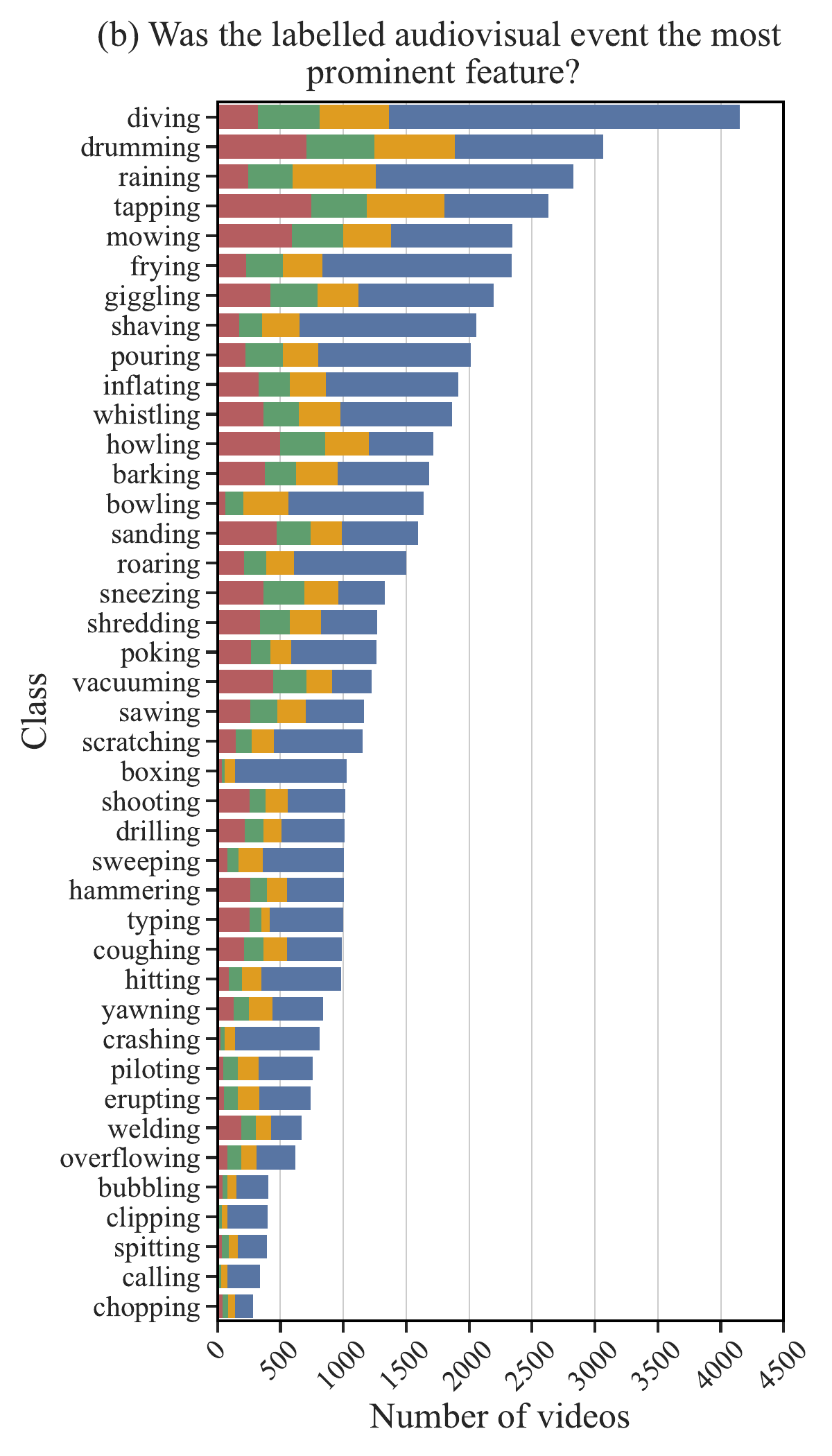}
    \end{subfigure}

    \caption{Number of MIT videos in each class that obtained a `yes' vote from 0,1,2 or 3 participants when asked the following questions: (a) Was the labelled audiovisual event present? (b) Was the labelled audiovisual event the most prominent feature?}

    \label{fig:perceived and dominant AVMIT}
\end{figure}

The AVMIT annotations are available in the file named video\_ratings.csv. Each row in the csv file corresponds to a video (containing all corresponding ratings from participants). Each video was rated 3 times. Videos rated less than 3 times were removed. The video\_ratings.csv fields are described in Table \ref{tab:video ratings csv}. The annotations are visualised in Figure \ref{fig:perceived and dominant AVMIT}. The test set details are provided in test\_set.csv, fields are described in Table \ref{tab:test set csv}.

\begin{table*}[h]
\begin{center}
\caption{Description of data in video\_ratings.csv}
\label{tab:video ratings csv}
\begin{tabular}{|c|c|}
    \hline
    Field & Description \\
    \hline
    filename & "MIT class subdirectory/ video name" \\
    \hline
    r1 & number of '1' ratings given \\
    \hline
    r2 & number of '2' ratings given \\
    \hline
    r3 & number of '3' ratings given \\
    \hline
    AVMIT\_label & as displayed to participants in annotation task \\
    \hline
    MIT\_label &	original dataset label \\
    \hline
    video\_location & training or validation directories of MIT \\
    \hline
    tfrecord\_filename & subdirectory and filename of corresponding audiovisual feature embeddings \\
    \hline
\end{tabular}
\end{center}
\end{table*}

\begin{table*}[h]
\begin{center}
\caption{Description of data in test\_set.csv}
\label{tab:test set csv}
\begin{tabular}{|c|c|}
    \hline
    Field & Description \\
    \hline
    filename & "MIT class subdirectory/ video name" \\
    \hline
    AVMIT\_label & as displayed to participants in annotation task \\
    \hline
    MIT\_label &	original dataset label \\
    \hline
    video\_location & training or validation directories of MIT \\
    \hline
    new\_filename & "AVMIT label subdirectory/ new video name" \\
    \hline
    tfrecord\_filename & subdirectory and filename of corresponding audiovisual feature embeddings \\
    \hline
\end{tabular}
\end{center}
\end{table*}

There are 2 archived feature embedding directories; AVMIT\_VGGish\_VGG16.tar contains the audiovisual embeddings, extracted by VGGish (audio) and VGG-16 (visual) for all AVMIT videos, AVMIT\_YamNet\_EffNetB0.tar contains the audiovisual embeddings extracted by YamNet (audio) and EfficientNetB0 (visual) for all AVMIT videos. Both sets of feature embeddings have the same directory structure, containing 1 subdirectory per action class (e.g. `barking') for all 41 classes. Inside each class sub-directory lies a .tfrecord file for each AVMIT video. Each tfrecord contains a number of context features; filename, label, number of audio timesteps, number of visual timesteps and 2 sequence features; audio data and visual data. For YamNet-EffNetB0 embeddings, audio data has dimensions (timesteps, 1,024) and visual data has dimensions (timesteps, 1,280). For VGGish-VGG16 embeddings, audio data has dimensions (timesteps, 128) and visual data has dimensions (timesteps, 512).

\section*{Usage Notes}
AVMIT is available at \url{https://zenodo.org/record/8253350}. To use the audiovisual feature embeddings, provided as part of this work, directly. An example python script, feature\_extractor/read\_tfrecords.py, is provided at \url{https://github.com/mjoannou/audiovisual-moments-in-time} to demonstrate how to read these tfrecords into a tensorflow.data.Dataset. AVMIT annotations in video\_ratings.csv can be used to filter these embeddings for audiovisual content, and test\_set.csv can be used to identify those embeddings intended for testing.

To use raw videos, one needs to download the well-established Moments in Time dataset by visiting \url{http://moments.csail.mit.edu/} and fill out a form before access to the dataset is sent via email. Once access to the MIT dataset is granted, AVMIT annotations, available in video\_ratings.csv, can be used to filter videos according to audiovisual content prior to training computational models. The AVMIT test set can also be used alongside the MIT videos, identifying 960 videos suitable for testing computation models and human participants alike. If one wishes to extract tfrecords, in a similar manner to our work, this is demonstrated in feature\_extractor/extract\_features.py.

\section*{Experiment}

\subsection*{Outline}
We sought to both validate the DNN embeddings and demonstrate the utility of the AVMIT annotations beyond the existing MIT annotations. To do this, we used the AVMIT audiovisual labels to filter embeddings and train a series of DNNs exclusively on audiovisual actions. We then tested these DNN classifiers vs. DNNs trained on a much larger set of embeddings sampled from MIT classes. The expected utility of the AVMIT embeddings was improved performance in the audiovisual domain, and so the AVMIT test set was used as a controlled audiovisual benchmark.

\subsection*{Train Sets}
Two train sets were prepared; an audiovisual train set using AVMIT annotations and a larger modality-agnostic (audio and/or visual) train set of MIT embeddings named MIT-16. Both train sets contained embeddings corresponding to the 16 AVMIT test set classes. To prepare the audiovisual train set using AVMIT annotations, we obtained only those embeddings that contained the labelled audiovisual event as a prominent feature, according to majority participant vote. To construct the second train set, all MIT embeddings corresponding to the 16 AVMIT classes were obtained. Embeddings corresponding to the AVMIT test were then removed from both train sets. Finally, the number of train embeddings across each class was balanced by sampling the maximum possible number of embeddings (AVMIT: 456 per class, MIT-16: 1,406 per class).

\subsection*{DNN Architectures}
Each architecture effectively consisted of a (frozen) AudioSet-trained CNN, a (frozen) ImageNet-trained CNN, some shared (trainable) audiovisual operations followed by a (trainable) RNN. For the CNNs, architectures either used VGGish \cite{Hershey2017} (audio) and VGG-16 \cite{Simonyan2015} (visual) or YamNet \cite{YamNet} (audio) and EfficientNetB0 \cite{Tan2019} (visual). Although practically, we provide these embeddings as part of this work and we trained on them directly. These architectures allowed us to leverage powerful pretrained unimodal representations but ensure that any learnt audiovisual features would arise from training on AVMIT/MIT-16 alone. We select similar architectures in each set of embeddings to help prevent overpowered unimodal representations in the trained classifiers and ensure both auditory and visual embeddings are useful.

\begin{figure}[h]
  \centering
  \includegraphics[width=0.75\textwidth]{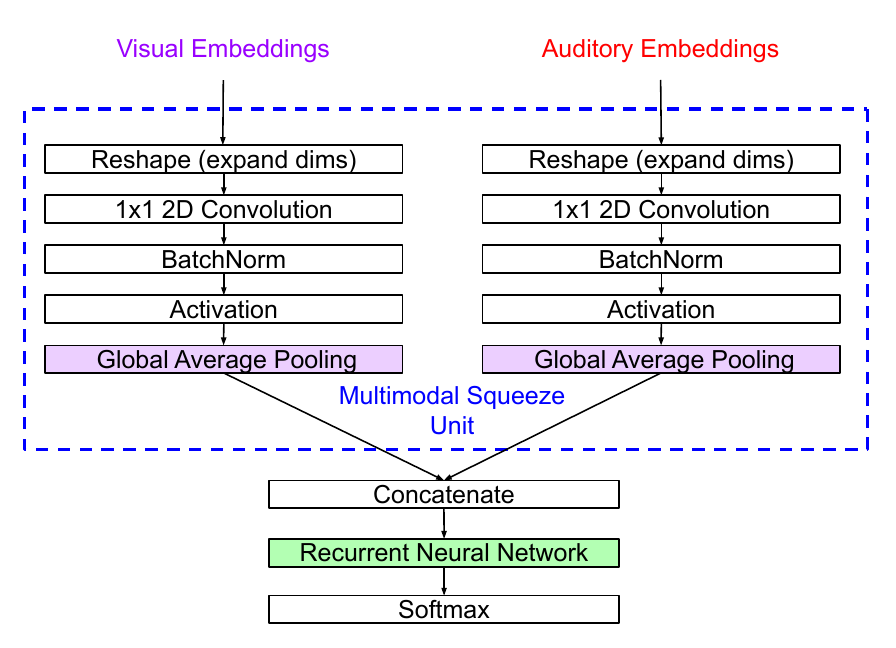}
\caption{Processing stream for audiovisual feature embeddings.} 
\label{fig:multimodal_squeeze_unit}
\end{figure}

As the audio and visual embeddings are of different sizes, we added batch-norm convolutional layers and global average pooling operations to each, individually, prior to concatenation. We refer to this series of processes as a multimodal squeeze unit (Figure \ref{fig:multimodal_squeeze_unit}). This is to ensure that there are an equal number of RNN connections dedicated to the processing of auditory and visual information. Following the multimodal squeeze unit, was one of three well-known RNN architectures; fully-recurrent neural network (FRNN, also known as a ‘basic’ or ‘vanilla’ RNN), gated recurrent unit \cite{ChoGRU2014} or a long short-term memory unit \cite{Hochreiter1997}.

% MULTIMODAL SQUEEZE UNIT

\subsection*{Hyperparameter Search and Training}
% Hyperparameter search
We ran a hyperparameter search (random search) on each embedding-set/RNN combination with the MIT-16 dataset. By optimising the hyperparameters on the MIT-16 dataset, we biased the experiment in favour of MIT-16 trained RNNs, thus strengthening any observed AVMIT related performance gains. For each embedding-set/RNN combination (2 x 3 = 6), we created 300 surrogate models, each with a particular combination of hyperparameter values that were uniformly sampled from the hyperparameter sets or intervals.

% Hyperparameter search space
We searched over the following hyperparameters; number of filters, $n_{bottleneck}$, in the audiovisual bottleneck ( 1x1 2D Convolution) where $n_{bottleneck} \in  \{32, 64, 128, 256\}$, the activation function, $a$,  of the audiovisual bottleneck, where $a \in \{relu, swish\}$, the number of Recurrent Neural Network units, $n_{RNN}$, where $n_{RNN} \in \{32, 64, 128, 256\}$, the dropout rate, $d$, for the RNN, where $d \in \{0.1, 0.2, 0.3, 0.4, 0.5\}$ and the learning rate, $l$, of the model, where $l \in [1.0 x 10\textsuperscript{-5}, 5.0 x 10\textsuperscript{-3}]$. During the random search, RNNs were trained in the same manner (Adam optimiser \cite{Kingma2015} and exponential learning rate decay) as during final training, the only exception being that early stopping was reduced from 20 epochs to 8 in order to save time during the random search. The best performing configurations for each RNN (Table \ref{tab:hyperparameter search}) were selected for comparison across all experiments.

\begin{table*}[h]
\begin{center}
\caption{Hyperparameter Search Results: Selected Hyperparameters} \label{tab:hyperparameter search}
\begin{tabular}{|c|c|c|c|c|c|c|c|}
  \hline
  Embeddings & RNN & \makecell{RNN \\ Units} & \makecell{Bottleneck \\ Units} & Activation & \makecell{Dropout \\ Rate} & \makecell{Learning \\ Rate} & \makecell{Trainable \\ Params}
  \\
  \hline
  YamNet + EffNetB0 & FRNN & 128 & 256 & swish & 0.3 & $7.05 x 10\textsuperscript{-5}$ & 675,472 \\
  \hline
  YamNet + EffNetB0 & GRU & 128 & 64 & swish & 0.5 & $7.25 x 10\textsuperscript{-5}$ & 248,976 \\
  \hline
  YamNet + EffNetB0 & LSTM & 64 & 256 & swish & 0.3 & $4.10 x 10\textsuperscript{-5}$ & 740,112 \\
  \hline
  VGGish + VGG-16 & FRNN & 256 & 256 & swish & 0.4 & $1.05 x 10\textsuperscript{-4}$ & 366,352 \\
  \hline
  VGGish + VGG-16 & GRU & 128 & 256 & relu & 0.5 & $3.92 x 10\textsuperscript{-4}$ & 413,968 \\
  \hline
  VGGish + VGG-16 & LSTM & 256 & 256 & swish & 0.5 & $1.74 x 10\textsuperscript{-4}$ & 956,944 \\
  \hline
\end{tabular}
\end{center}
\end{table*}

For each hyperparameter combination, we trained one RNN instance (row in Table \ref{tab:hyperparameter search}) on AVMIT, and another instance on MIT-16. The cross-entropy loss function was used as a measure of loss, and the RNN was trained with backpropagation and the Adam optimiser \cite{Kingma2015}. Each RNN was trained for up to 200 epochs with a batch size of 16 samples, although with an early stopping of 20 epochs, all RNNs executed training before that point. All learned parameters were then fixed in place throughout testing.

\subsection*{Evaluation Method}
The AVMIT controlled test set was used for testing. As the test set had been well filtered to include only prominent audiovisual events, any learnt audiovisual features should be beneficial to performance. The loss, top 1 classification accuracy (the proportion of trials in which the model gave the highest probability to the correct action class) and the top 5 classification accuracy (the proportion of trials in which the correct action class was assigned one of the top five probabilities) was used to measure performance on this set.

% Results
\subsection*{Results}
All models obtained a top 5 classification accuracy of approximately 100\%. Models trained on AVMIT obtained a lower loss and higher top 1 accuracy than their MIT-16 trained counterpart in all cases (Table \ref{tab:action recognition results}). This result indicates that training a DNN exclusively on audiovisual action events is beneficial for audiovisual action recognition, even outweighing a three-fold increase in training data (additional audio or visual events). A final observation is that the YamNet+EfficientNet-B0 embeddings consistently provided higher performances than VGGish+VGG-16 embeddings.

% AUDIOVISUAL PERFORMANCE TABLE
\begin{table*}[h]
\begin{center}
\caption{Action Recognition Performance on AVMIT Test Set.} \label{tab:action recognition results}
\begin{tabular}{|c|c|c|c|c|c|}
  \hline
  Training Set & Embeddings & RNN & Loss & Top 1 & Top 5\\ & & & & Acc. (\%) & Acc. (\%)
  \\
  \hline
  AVMIT & YamNet + EffNetB0 & FRNN & 0.1841 & 94.58 & 99.90 \\
  MIT 16 & YamNet + EffNetB0 & FRNN & 0.2973 & 89.79 & 99.90 \\
  \hline
  AVMIT & YamNet + EffNetB0 & GRU & 0.1600 & 95.73 & 99.90 \\
  MIT 16 & YamNet + EffNetB0 & GRU & 0.2430 & 92.29 & 99.90 \\
  \hline
  AVMIT & YamNet + EffNetB0 & LSTM & 0.1674 & 95.52 & 99.79 \\
  MIT 16 & YamNet + EffNetB0 & LSTM & 0.2366 & 92.81 & 100 \\
  \hline
  AVMIT & VGGish + VGG-16 & FRNN & 0.2980 & 90.73 & 99.79 \\
  MIT 16 & VGGish + VGG-16 & FRNN & 0.4388 & 84.79 & 99.58 \\
  \hline
  AVMIT & VGGish + VGG-16 & GRU & 0.2917 & 91.04 & 99.79 \\
  MIT 16 & VGGish + VGG-16 & GRU & 0.4108 & 85.83 & 99.69 \\
  \hline
  AVMIT & VGGish + VGG-16 & LSTM & 0.2892 & 90.94 & 99.90 \\
  MIT 16 & VGGish + VGG-16 & LSTM & 0.3527 & 86.98 & 99.90 \\
  \hline
\end{tabular}
\end{center}
\end{table*}

\section*{Conclusion}
We present Audiovisual Moments in Time, a set of audiovisual annotations and DNN embeddings for the Moments in Time dataset. AVMIT contains annotations of 57,177 videos across 41 classes, each pertaining to the existence of an audiovisual event, and its prominence in the video. We demonstrate the utility of AVMIT audiovisual annotations beyond unimodal annotations by training a series of RNNs exclusively on audiovisual data vs. modality-agnostic (audio and/or visual) data and observing an increase of 2.71-5.94\% in top 1 accuracy on our audiovisual test set. We further provide a set of 960 videos (60 videos over 16 classes), designated as a controlled test set. These videos can be manipulated for audiovisual synchrony, semantic correspondence, visual or auditory noise etc. to produce a large suite of test videos, suitable for experiments with DNNs and humans alike. Finally, we provide DNN embeddings for AVMIT videos to lower the computational barriers for those who wish to train audiovisual DNNs, thereby levelling the playing field for all. AVMIT provides a useful resource for experiments concerned with audiovisual correspondence, and allows DNN comparisons against humans to take a step into the audiovisual domain.

\section*{Acknowledgement}
We would like to thank the Engineering and Physical Sciences Research Council (ESPRC) and the ERC (ERC starting grant: multsens) for funding this research.

\section*{Author contributions statement}
M.J. carried out all of the work in this research paper under the supervision of P.R. and U.N. All authors contributed to manuscript drafts and gave final approval for publication.

\bibliographystyle{ieeetr}
\bibliography{references}

\end{document}